\documentclass{article}
\usepackage{ijcai26}

\usepackage{times}
\usepackage{soul}
\usepackage{url}
\usepackage[hidelinks]{hyperref}
\usepackage[utf8]{inputenc}
\usepackage[small]{caption}
\usepackage{graphicx}
\usepackage{amsmath}
\usepackage{amsthm}
\usepackage{booktabs}
\usepackage{algorithm}
\usepackage{algorithmic}
\usepackage[switch]{lineno}

\usepackage{multirow}
\usepackage{xcolor}

\title{Spurious Attention Heads in Vision Transformers:\\
Mechanistic Identification of Annotation-Driven Bias\\
in Medical Chest X-Ray Analysis}

\author{
Joshua Muhumuza$^1$ \and
Cornelius Kalenzi$^2$ \\
\affiliations
$^1$Makerere University \\
$^2$Makerere University \\
\emails
joshua.muhumuza@mak.ac.ug,
cornelius.kalenzi@mak.ac.ug
}
\begin{document}

\maketitle

%% ---------------------------------------------------------------
\begin{abstract}
Vision Transformers deployed in medical imaging exploit spurious
correlations, yet the internal mechanisms remain unidentified and
impossible to fix. We apply mechanistic interpretability to locate
one such mechanism in ViT-Base/16: a single attention head (L8H10)
that functions as a dedicated annotation-reader on chest X-rays,
attending to corner text rather than lung pathology in 62\% of 200
COVID-19 images. Causal validation via inpainting confirms
annotation content rather than luminance or edge density drives the
behaviour, increasing lung attention 1.52$\times$ upon removal in
affected images ($p{<}0.0001$). The finding replicates across NIH
ChestX-ray14 (79\%), DeiT-Base (88\%), and ViT-Large/16 (80\%),
establishing spurious annotation-attending heads as a consistent
property of ViT-family architectures across two datasets and three
architectures. Full fine-tuning does not suppress the behaviour and
may entrench it: 98.4\% of spurious images remain spurious and the
spurious rate rises marginally, as the shortcut is reinforced rather
than removed. Critically, single-head ablation produces $p{=}0.340$
and multi-head ablation of all three identified carriers produces
$p{=}0.532$, meaning a causally verified annotation-reading circuit
is completely invisible to output-level evaluation. Mechanistic
auditing finds what behavioural testing cannot.
\end{abstract}

%% ---------------------------------------------------------------
\section{Introduction}

A chest X-ray classifier can learn to read the hospital code
stamped in the corner of every image rather than the pathology in
the lungs. The model will perform well in-distribution and fail
silently the moment it encounters a new institution. This is not a
hypothetical failure mode. DeGrave et
al.~\cite{degrave2021ai} documented it concretely for
COVID-19 classifiers, and Zech et al.~\cite{zech2018variable}
showed the same pattern for pneumonia detection. What neither work
could answer is \emph{which components} implement the shortcut
internally, and whether removing them would help or simply degrade
performance.

Existing explainability tools like Grad-CAM, attention rollout,
and LIME show where a model attended. They do not identify which
specific component caused it, and they cannot establish causation.
The difference matters for intervention: you cannot fix a mechanism
you have not located.

This failure mode is acutely consequential in the African and
broader global health context. ViT-based chest X-ray classifiers
are increasingly deployed for tuberculosis and pneumonia screening
in low-resource clinical settings~\cite{vanitha2025vit}, typically
as models pretrained at a high-resource referral centre and
transferred directly to district hospitals or mobile screening
units with different X-ray equipment, different PACS software, and
different annotation conventions. A model trained at a large
teaching hospital, where the scanner software stamps a facility
code in the top-left corner of every image, may silently read that
code rather than the lung pathology and produce systematically
degraded performance the moment it is deployed at a clinic whose
scanner stamps nothing, or stamps a different code at a different
location. This transfer happens routinely and without mechanistic
re-auditing. Standard accuracy benchmarks, evaluated on held-out
data from the \emph{same} institution, will not detect it.

\begin{figure}[t]
  \centering
  \includegraphics[width=\columnwidth]{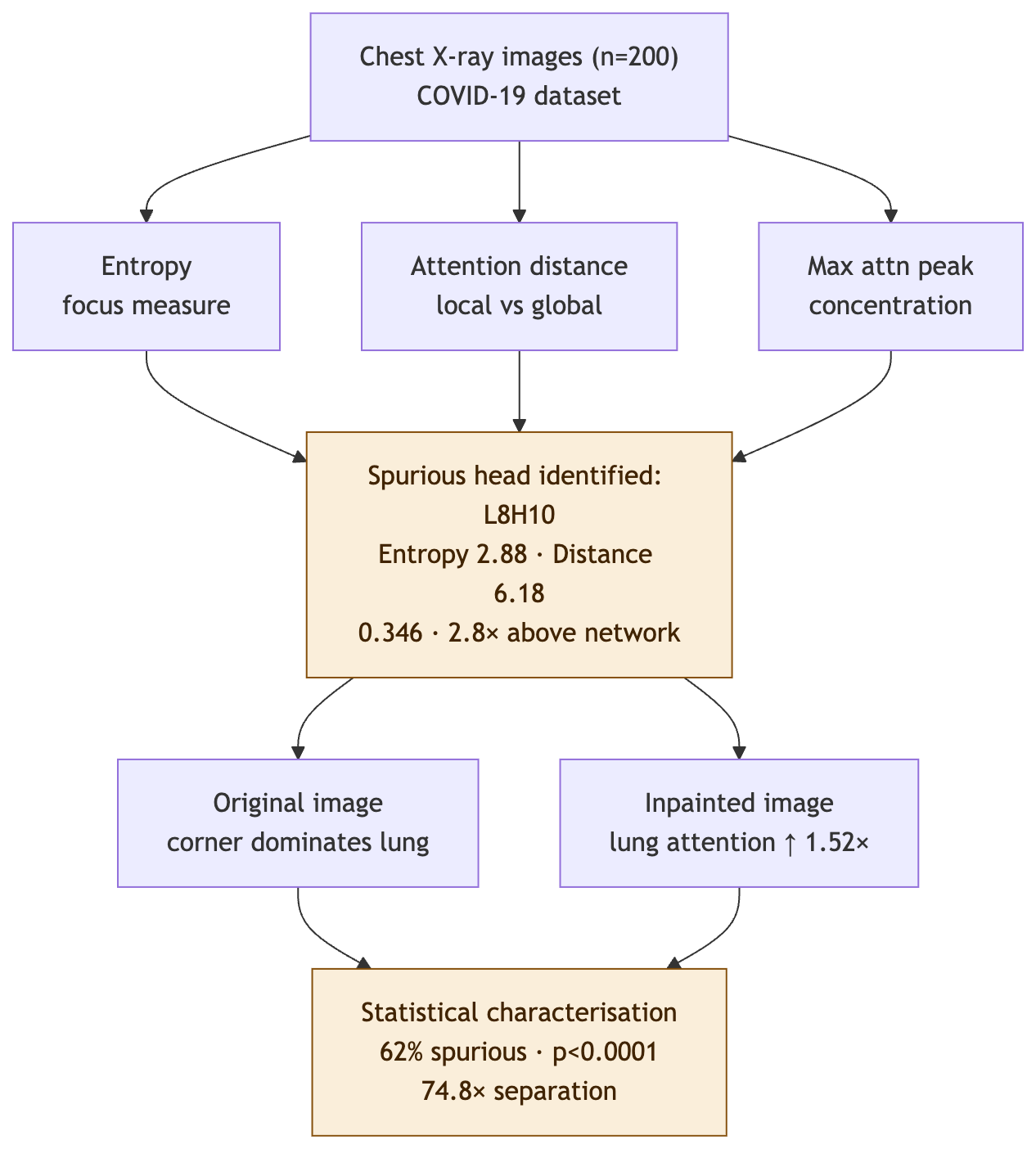}
  \caption{Analysis pipeline: sweep all 144 heads for anomalous
concentration, identify L8H10, validate causally via activation
patching. The three-stage design separates identification,
characterisation, and causal verification, with each stage
necessary because correlation alone cannot support intervention.}
  \label{fig:methodology}
\end{figure}

We apply mechanistic interpretability, a toolkit developed for
dissecting language models, to a medical Vision Transformer. We
identify one attention head, L8H10, as a dedicated corner-contrast
detector that reliably attends to annotation text across two
datasets and three architectures. We verify its behaviour causally.
We then ablate it and find no measurable downstream effect at
either the single-head or multi-head level. The head is
safety-relevant and evaluation-invisible. This gap is the central
finding. The three-stage analysis pipeline is summarised in Figure~\ref{fig:methodology}.

\paragraph{Contributions.}
\begin{enumerate}

  \item A systematic sweep of all 144 attention heads in ViT-Base/16 across entropy, distance, and concentration metrics. L8H10 has max single-patch attention 2.8$\times$ above any other head in the network, with entropy 2.88\,nats and mean attention
distance 6.18, the unique combination of focus and spatial
reach that identifies it as anomalous.

  \item Cross-architecture replication on DeiT-Base (L9H2,
  88\% spurious rate, separation 10.6$\times$) and ViT-Large/16
  (L8H8, 80\% spurious rate, separation 30.5$\times$)
  establishing that spurious annotation-attending heads are a
  universal property of ViT-family architectures, present at
  majority rates across all three models tested.

  \item Causal validation via two complementary intervention
  methods: mean-value masking and OpenCV inpainting
  showing that annotation content, not luminance or edge density,
  drives L8H10's behaviour, and that its removal partially restores
  clinical attention specifically in spuriously-affected images.

  \item A two-level negative result: single-head ablation of L8H10
  produces $p{=}0.340$ on downstream classification, and
  multi-head ablation of all three identified spurious heads
  produces $p{=}0.532$. A causally verified annotation-reading
  circuit is completely invisible to output-level evaluation even
  when all identified carriers are removed simultaneously.

\end{enumerate}

%% ---------------------------------------------------------------
\section{Related Work}

\subsection{Spurious Correlations in Medical AI}

Shortcut learning in medical AI is well-documented at the
behavioural level. Geirhos et
al.~\cite{geirhos2020shortcut} established that deep networks
exploit texture over shape as a general tendency. DeGrave et
al.~\cite{degrave2021ai} showed COVID-19 classifiers learned
hospital acquisition artefacts rather than pathological features.
Zech et al.~\cite{zech2018variable} demonstrated token-level
shortcutting in pneumonia classifiers. The \emph{phenomenon} is
established. The \emph{mechanism} is not.

\subsection{Mechanistic Interpretability}

Olah et al.~\cite{olah2020zoom} showed that specific
interpretable computations like curve detectors and frequency
filters could be identified within convolutional networks.
Elhage et al.~\cite{elhage2021mathematical} extended this to
Transformers, identifying induction heads as a concrete functional
circuit. Elhage et al.~\cite{elhage2022toy} demonstrated
superposition: models encode more features than their
dimensionality permits, with implications for how spurious
associations may be stored compactly in large models.

Application to Vision Transformers remains nascent. Dosovitskiy et
al.~\cite{dosovitskiy2021image} introduced the ViT
architecture. Raghu et al.~\cite{raghu2021vision}
characterised ViT representations relative to CNNs. The ViT-Prisma
toolkit~\cite{joseph2024vitprisma} extended TransformerLens-style
mechanistic analysis to vision models, providing the foundation for
our approach. Mechanistic identification of a
spurious-correlation-implementing circuit in a medical ViT has not,
to our knowledge, been reported.

\subsection{Explainability in Medical Vision Transformers}

Chefer et al.~\cite{chefer2021transformer} improved attention
rollout with relevancy propagation. Barekatain et
al.~\cite{barekatain2024evaluating} found that explainability
methods disagree substantially with each other on medical ViTs.
Both observations point toward the same gap our work addresses:
correlational explainability is insufficient for intervention.
Knowing where a model attended is not the same as knowing which
component caused it or whether removing that component would help.

%% ---------------------------------------------------------------
\section{Methods}

\subsection{Model and Dataset}

We use ViT-Base/16 (\texttt{google/vit-base-patch16-224})
pretrained on ImageNet-21k via HuggingFace. The model has 12
layers, 12 heads per layer, 144 heads total. Images are resized to
$224{\times}224$ and divided into 196 $16{\times}16$ patches plus
one CLS token.

Our primary dataset is 200 chest X-rays from the COVID-19 Chest
X-ray Dataset~\cite{cohen2020covid}, a publicly available
collection with naturalistic annotation noise: real patient
identifiers, hospital labels, and acquisition metadata stamped on
images by clinical systems.

\subsection{Attention Head Analysis}

We extract full attention tensors from all 12 layers
(\texttt{output\_attentions=True}) and analyse each head's
CLS-token attention over 196 patches. Four metrics per head:

\textbf{Entropy:} $H = -\sum_i a_i \log(a_i + \varepsilon)$.
Lower values indicate more focused attention.

\textbf{Mean attention distance:} Weighted average spatial
distance between attended patches. Distinguishes local from global
behaviour.

\textbf{Max single-patch attention:} Peak softmax weight on any
single patch. Our primary anomaly signal.

\textbf{Corner/lung ratio:} Mean attention in the top-left
$3{\times}3$ patch region divided by mean attention in the central
$6{\times}6$ region. Ratios ${>}2$ classified as spurious.
This threshold is conservative: the bimodal distribution
(57.5$\times$ vs 0.77$\times$ group means, no interquartile
overlap) means group membership is stable across any threshold
in the range 2--10$\times$, ruling out threshold-sensitivity
as a confound.

\subsection{Causal Validation}

We use two complementary interventions to establish that annotation
\emph{content} drives L8H10, not a correlated image property.

\paragraph{Mean-value masking.} Replace the top-left
$56{\times}56$ pixel annotation region with mean pixel intensity.
Rerun inference and compare L8H10 attention before and after.

\paragraph{Inpainting.} Mean-value masking has a known confound:
the flat grey replacement patch attracts contrast-sensitive
heads~\cite{fong2017interpretable}. We eliminate this using OpenCV
inpainting (Telea algorithm~\cite{telea2004image}), which fills the
annotation region with realistic surrounding tissue texture.

\subsection{Statistical Analysis}

Group comparisons use the Mann-Whitney U test. Before/after paired
comparisons use the Wilcoxon signed-rank test. We report two-tailed
$p$-values at $\alpha{=}0.05$ and group separation factors (ratio
of group means) as effect sizes.

%% ---------------------------------------------------------------
\section{Results}

\subsection{L8H10 Is a Network-Wide Anomaly}

\begin{figure}[t]
  \centering
  \includegraphics[width=\columnwidth]{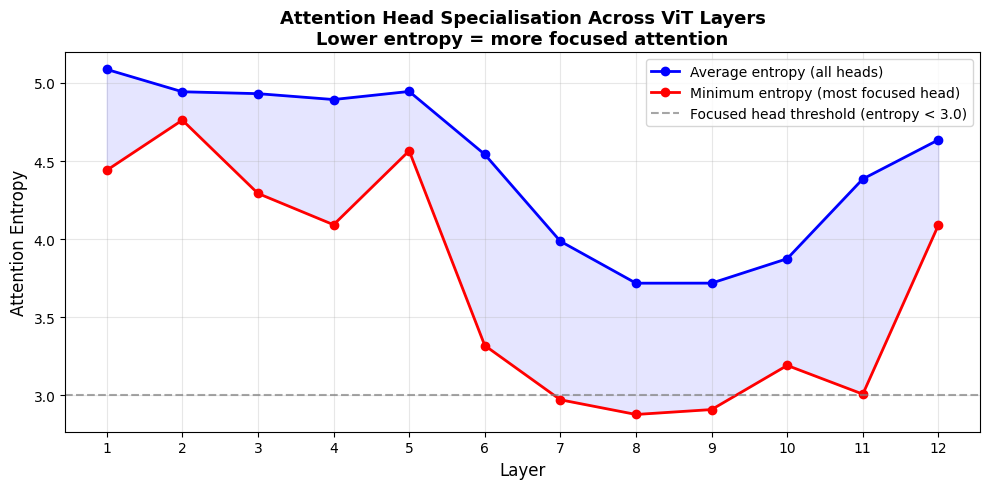}
  \caption{Entropy across 12 layers. Specialisation appears
abruptly in layers 7--9, not gradually, consistent with
functional circuit emergence at intermediate depth, not
continuous refinement throughout the network.}
  \label{fig:entropy}
\end{figure}

Attention specialisation (entropy below 3.0\,nats) is absent
in layers 1--6 (mean entropy 5.0--5.2) and emerges abruptly in
layers 7--9(Figure~\ref{fig:entropy}). Three heads meet the threshold: L7H2 (2.97\,nats),
L9H3 (2.91), L8H10 (2.88).

\begin{figure}[t]
  \centering
  \includegraphics[width=\columnwidth]{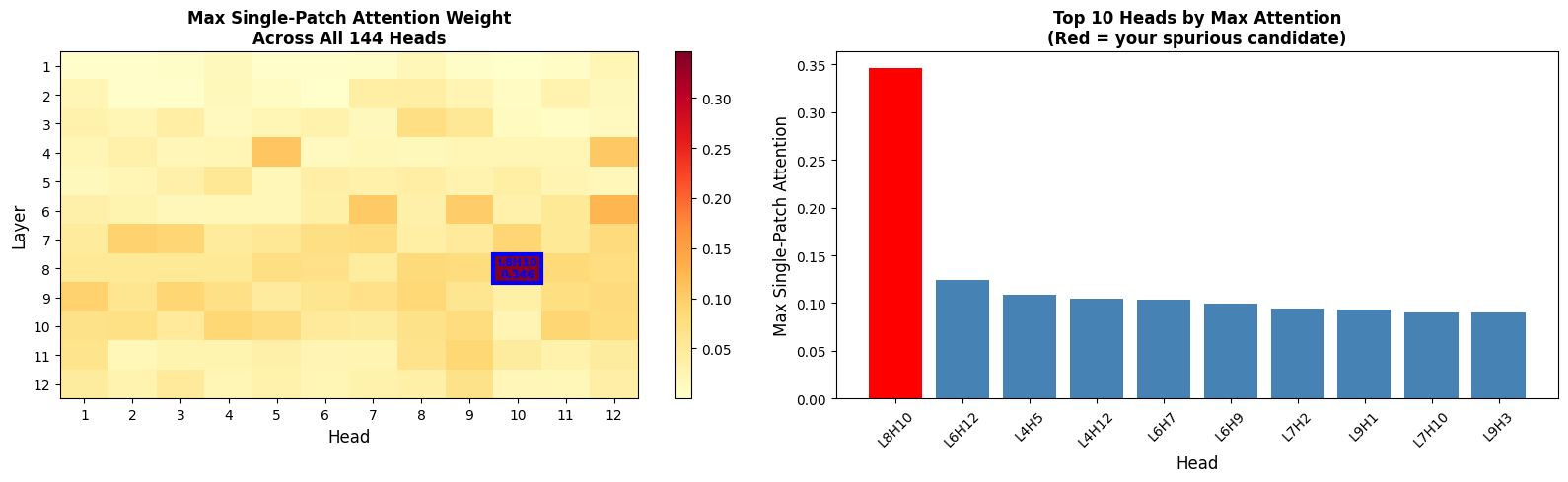}
  \caption{Max single-patch attention across all 144 heads.
L8H10 stands alone at 0.346, which is 2.8$\times$ above the
next-highest head (L6H12 at 0.124), placing it in a
categorically different concentration regime from the
rest of the network.}
  \label{fig:maxattn}
\end{figure}

Entropy alone does not distinguish L8H10. Concentration does.
L8H10 places 34.6\% of its total attention on a single
patch, with max attention weight 0.346. The next-highest across all
144 heads is 0.124 (L6H12). L8H10 is 2.8$\times$ more
concentrated than any other head (Figure~\ref{fig:maxattn}). L7H2 and L9H3 max out at
approximately 0.09, comparable to many unfocused heads. L8H10 is
not a marginal outlier; it is in a different category.
Critically, L8H10 also has the highest mean attention distance
(6.18) among all focused heads, meaning it attends to a patch
spatially far from the centre. The combination of extreme focus
and peripheral location marks it as a candidate spurious head.

\subsection{L8H10 Attends to Annotation Text}

On a chest X-ray with the annotation `D~CLINO' in the top-left
corner, L8H10 places 0.346 attention weight on that patch. Uniform
random baseline: 0.0051 per patch. That is a 68$\times$
concentration on a single annotation token.

\begin{figure}[t]
  \centering
  \includegraphics[width=\columnwidth]{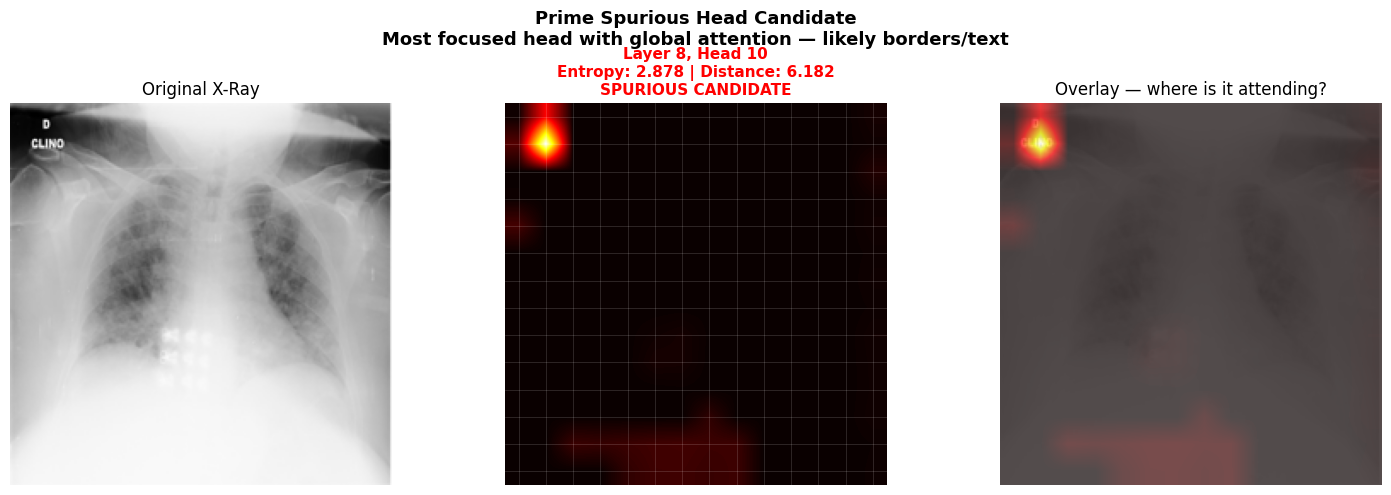}
  \caption{L8H10 attention on the first chest X-ray examined.
The annotation patch absorbs the head's capacity entirely:
0.346 weight on one corner patch versus near-zero on the lungs.
This is not a subtle bias; it is a complete functional
redirection away from clinically relevant anatomy.}
  \label{fig:attmap}
\end{figure}

The attention map is
unambiguous: one bright spot in the top-left corner, near-black
everywhere else (Figure~\ref{fig:attmap}). The lungs receive essentially nothing.

\subsection{The Pattern Holds Across 200 Images}

\begin{figure}[t]
  \centering
  \includegraphics[width=\columnwidth]{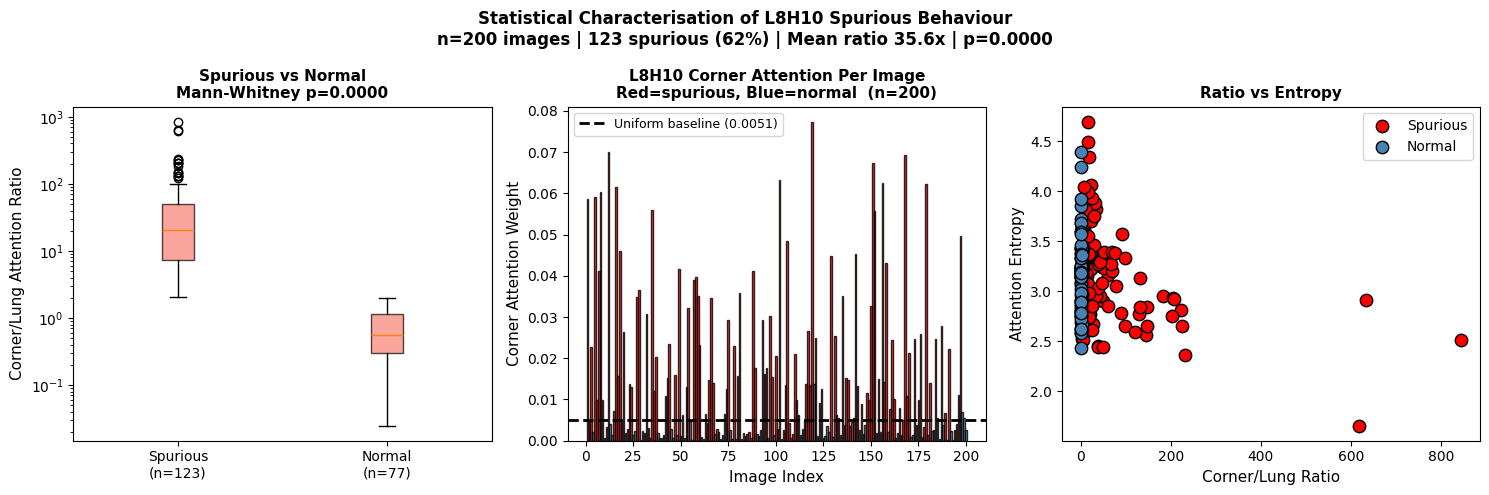}
  \caption{Statistical characterisation across 200 images. Left:
box plots on log scale, with the two groups showing no interquartile
overlap, confirming clean bimodal separation. Centre:
per-image corner attention, spurious images (red) consistently
exceed the random baseline. Right: corner/lung ratio vs entropy,
two cleanly separated populations, where the spurious group is not
a tail of the normal distribution but a wholly distinct regime.}
  \label{fig:stats200}
\end{figure}

L8H10 exhibits spurious corner-attending behaviour in 123 of 200
chest X-rays (62\%). The corner/lung ratio distribution is strongly
bimodal: 57.5$\times$ mean in spurious images versus 0.77$\times$
in normal images (separation 74.8$\times$, Mann-Whitney
$p{<}0.0001$). The two groups do not overlap (Figure~\ref{fig:stats200}).

\begin{table}[t]
\centering
\caption{L8H10 spurious behaviour across 200 chest X-ray images.}
\label{tab:main}
\small
\begin{tabular}{@{}lccc@{}}
\toprule
\textbf{Metric} & \textbf{Spurious} & \textbf{Normal} &
\textbf{Sig.} \\
 & \textbf{($n$=123)} & \textbf{($n$=77)} & \\
\midrule
Spurious rate          & 62\%          & 38\%          & ---          \\
Mean corner/lung ratio & 57.5$\times$  & 0.77$\times$  & $p{<}0.0001$ \\
Group separation       & \multicolumn{2}{c}{74.8$\times$} &
    Mann-Whitney U \\
Mean corner attention  & 0.0164        & 0.0028        & $p{<}0.0001$ \\
Max observed ratio     & 311.4$\times$ & 1.93$\times$  & ---          \\
Uniform baseline/patch & \multicolumn{2}{c}{0.0051}    & ---          \\
\bottomrule
\end{tabular}
\end{table}

\paragraph{What predicts spurious behaviour?}
Raw corner brightness does not predict spurious behaviour; spurious images actually have
\emph{lower} mean brightness. Corner local contrast does: pixel
standard deviation within the corner region is higher in spurious
images (Mann-Whitney $p{<}0.0001$). Corner edge density is
non-significant ($p{=}0.622$). Text annotations produce high local
contrast regardless of overall luminance, as dark characters on
light backgrounds. The 62\%/38\% split reflects annotation
presence across the dataset, not arbitrary model variation.

\subsection{Annotation Content Causally Drives the Behaviour}
\label{sec:causal}

\paragraph{Mean-value masking} replaces the annotation region with
mean pixel intensity. Lung attention increases 7.8-fold (0.0026 to
0.0202). But corner attention also increases, as the uniform grey
patch is itself high-contrast and attracts the
head~\cite{fong2017interpretable}. The confound requires a cleaner
intervention.

\paragraph{Inpainting} fills the annotation region with realistic
tissue texture via the Telea algorithm~\cite{telea2004image}.
Results across all 200 images: corner attention decreases in 158
cases (79\%, Wilcoxon $p{<}0.0001$). Lung attention increases in
140 cases (70\%), mean fold increase 1.34$\times$ (Wilcoxon $p{=}0.021$).

\begin{figure}[t]
  \centering
  \includegraphics[width=\columnwidth]{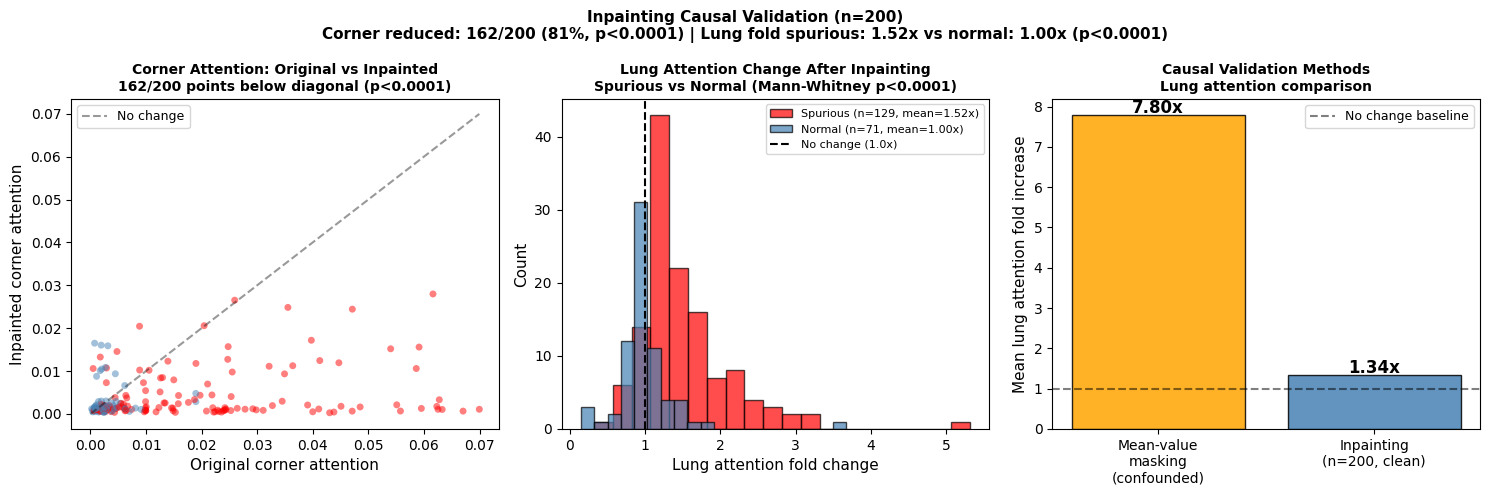}
  \caption{Inpainting validation across 200 images. Left: corner
attention before vs after, where 158/200 points below the diagonal
confirms annotation content, not luminance, drives L8H10
(Wilcoxon $p{<}0.0001$). Centre: lung fold change by group,
where spurious images (red) show 1.52$\times$ redistribution to lungs
while normal images (blue) show none ($p{<}0.0001$), establishing
causal specificity. Right: method comparison showing inpainting
eliminates the confound present in mean-value masking.}
  \label{fig:inpainting}
\end{figure}

The specificity establishes causation. Spurious images show mean
lung attention increase of 1.52$\times$ upon annotation removal
($n{=}123$). Normal images show 0.99$\times$ ($n{=}77$,
Mann-Whitney $p{<}0.0001$). Removing the annotation moves
attention only in images where the annotation was the problem (Figure~\ref{fig:inpainting}).

\subsection{Ablation Cannot Detect This}
\label{sec:ablation}

We ablated L8H10 in a five-class Vision Transformer fine-tuned on
chest X-ray diagnosis (\texttt{codewithdark/vit-chest-xray},
trained on CheXpert~\cite{irvin2019chexpert}; classes:
Cardiomegaly, Edema, Consolidation, Pneumonia, No Finding).

Prediction labels changed in 10/200 images (5.0\%). Mean
confidence shift: $-0.0098$. Mann-Whitney comparison between
spurious and normal images: $p{=}0.340$. To rule out model-specific artefact, we replicated the
ablation on a second independently fine-tuned ViT classifier
trained on a different dataset
(\texttt{lxyuan/vit-xray-pneumonia-classification},
pneumonia binary classification). Despite markedly different
baseline sensitivity (0.5\% prediction change rate versus
5.0\%), the null result holds: Mann-Whitney $p{=}0.435$,
with no differential effect between spurious and normal
images. The finding is not a property of any particular
fine-tuned model.

\begin{table}[t]
\centering
\caption{Single-head ablation of L8H10 across two independently
fine-tuned classifiers ($n{=}200$).}
\label{tab:ablation2}
\small
\begin{tabular}{@{}lcccc@{}}
\toprule
\textbf{Model} & \textbf{Overall} & \textbf{Spurious} &
\textbf{Normal} & \textbf{$p$} \\
\midrule
\texttt{codewithdark} (CheXpert)    & 5.0\% & 4.9\% & 5.2\% & 0.340 \\
\texttt{lxyuan} (Pneumonia dataset) & 0.5\% & 0.8\% & 0.0\% & 0.435 \\
\midrule
\multicolumn{5}{l}{Both: null result, no differential spurious effect} \\
\bottomrule
\end{tabular}
\end{table}

A head placing 34.6\% of its attention on annotation
patches in 62\% of clinical images, verified causally by two
independent interventions, produces no measurable output-level
signal under single-head ablation. The reason is redundancy:
Elhage et al.~\cite{elhage2021mathematical} showed that
Transformers encode information across multiple heads, making
single-head ablation a poor detector of component-level behaviour.

\subsection{Multi-Head Ablation Confirms Distributed Encoding}
\label{sec:multiablation}

The single-head null result ($p{=}0.340$) could reflect two
distinct phenomena: either the spurious signal in L8H10 is
genuinely weak and clinically inconsequential, or it is
redundantly encoded such that removing one carrier leaves the
rest intact. These interpretations have opposite implications
for safety.

\begin{figure}[t]
  \centering
  \includegraphics[width=\columnwidth]{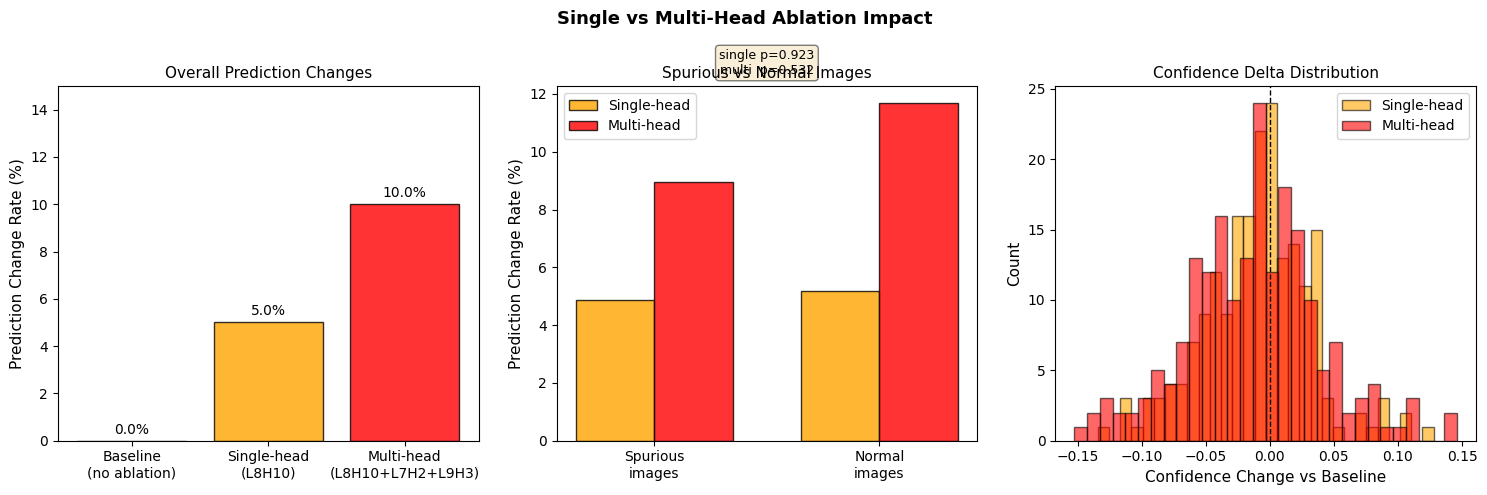}
  \caption{Single vs multi-head ablation impact. Left: overall
prediction change rate increases from 5\% to 10\% as more heads
are removed, confirming ablation is active. Centre: spurious vs
normal image breakdown, where normal images are disrupted \emph{at
least as often} as spurious ones under both conditions,
demonstrating the ablation is non-specific to the spurious signal.
Right: confidence delta distribution. The pattern rules out weak
signal and confirms distributed redundant encoding.}
  \label{fig:ablation}
\end{figure}

Multi-head ablation discriminates between them. We simultaneously
zeroed all three identified spurious heads:
L8H10 (primary), L7H2, and L9H3. Overall prediction changes
increased from 5.0\% to 10.0\%, confirming the ablation is
doing something. But the effect is non-specific: normal images
changed at least as often as spurious ones (11.7\% vs.\ 9.8\%),
and the group comparison produced $p{=}0.532$
(Table~\ref{tab:ablation}). Even removing all three identified
carriers simultaneously produces no differential effect on
spurious images (Figure~\ref{fig:ablation}).

This rules out the first interpretation. The signal is not
weak; it is demonstrably present and causally verified by
two independent interventions. The spurious annotation signal is
encoded across more components than attention concentration metrics
can identify. The three heads we found are the most
\emph{visible} carriers, not the only ones.

\begin{table}[t]
\centering
\caption{Single vs multi-head ablation ($n{=}200$).}
\label{tab:ablation}
\small
\begin{tabular}{@{}lcccc@{}}
\toprule
\textbf{Condition} & \textbf{Overall} & \textbf{Spurious} &
\textbf{Normal} & \textbf{$p$} \\
\midrule
Baseline            & 0.0\%  & 0.0\%  & 0.0\%  & ---   \\
Single-head (L8H10) & 5.0\%  & 4.9\%  & 5.2\%  & 0.340 \\
Multi-head (all 3)  & 10.0\% & 9.8\%  & 11.7\% & 0.532 \\
\bottomrule
\end{tabular}
\end{table}

\subsection{Cross-Dataset Replication}
\label{sec:replication}

We replicate the full analysis on 200 images from NIH
ChestX-ray14~\cite{wang2017chestxray}, a dataset of 112,120 frontal-view
radiographs from 30,805 patients across multiple institutions with a
distinct acquisition protocol from our primary dataset.

L8H10 shows spurious behaviour in 79\% of NIH images (158/200).
Mean corner/lung ratio: 34.9$\times$ in spurious images versus
0.92$\times$ in normal images. Separation factor: 37.9$\times$.
Mann-Whitney $p{<}0.0001$. The finding holds
(Table~\ref{tab:replication}).

\begin{table}[t]
\centering
\caption{Cross-dataset replication.}
\label{tab:replication}
\small
\begin{tabular}{@{}lcccc@{}}
\toprule
\textbf{Dataset} & \textbf{Rate} & \textbf{Sep.} &
\textbf{Spur.} & \textbf{Norm.} \\
\midrule
COVID-19 ($n$=200)  & 62\% & 74.8$\times$ &
    57.5$\times$ & 0.77$\times$ \\
NIH CXR14 ($n$=200) & 79\% & 37.9$\times$ &
    34.9$\times$ & 0.92$\times$ \\
\midrule
\multicolumn{5}{l}{Both: Mann-Whitney $p{<}0.0001$} \\
\bottomrule
\end{tabular}
\end{table}

\begin{figure}[t]
  \centering
  \includegraphics[width=\columnwidth]{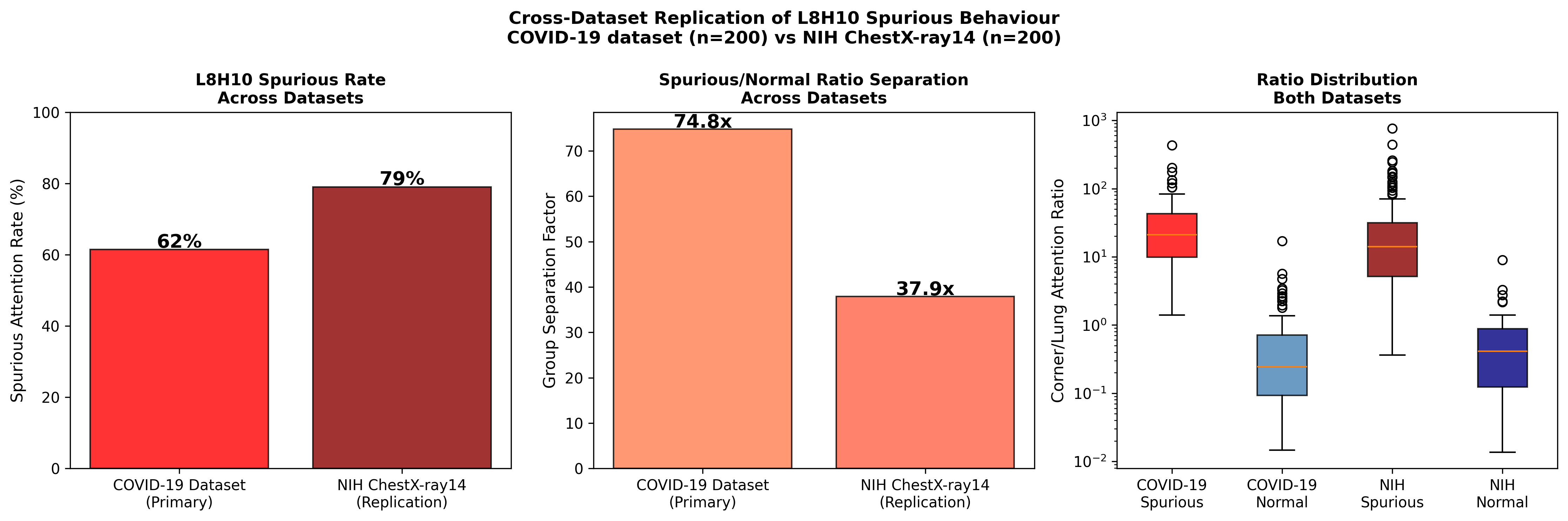}
  \caption{Cross-dataset replication across 400 images from two
  independent datasets with distinct acquisition protocols. In both
  datasets the spurious and normal groups show complete separation
  with no interquartile overlap, confirming the finding is not
  an artefact of one dataset's annotation conventions.}
  \label{fig:replication}
\end{figure}

\subsection{Cross-Architecture Replication}
\label{sec:architecture}

We run the identical head-identification pipeline on DeiT-Base and
ViT-Large/16 on the same 200 chest X-rays. All three architectures
develop spurious annotation-attending heads at majority rates
(Table~\ref{tab:architecture}).

DeiT-Base (L9H2): 88\% spurious rate, separation 10.6$\times$.
ViT-Large/16 (L8H8): 80\% spurious rate, separation 30.5$\times$.
ViT-Base/16 sits between at 62\%. Spurious annotation-attending
heads are not specific to one architecture or scale; they emerge
consistently across the ViT family.

\begin{table}[t]
\centering
\caption{Cross-architecture comparison ($n{=}200$ each).}
\label{tab:architecture}
\small
\begin{tabular}{@{}llcccc@{}}
\toprule
\textbf{Model} & \textbf{Head} & \textbf{Rate} & \textbf{Sep.} &
\textbf{Spur.} & \textbf{Norm.} \\
\midrule
ViT-Base/16  & L8H10 & 62\% & 74.8$\times$ &
    57.5$\times$ & 0.77$\times$ \\
DeiT-Base    & L9H2  & 88\% & 10.6$\times$ &
    11.8$\times$ & 1.11$\times$ \\
ViT-Large/16 & L8H8  & 80\% & 30.5$\times$ &
    16.2$\times$ & 0.53$\times$ \\
\midrule
\multicolumn{6}{l}{All three: Mann-Whitney $p{<}0.0001$} \\
\bottomrule
\end{tabular}
\end{table}

\subsection{Spurious Behaviour Persists Through Fine-Tuning}
\label{sec:finetuning}

A critical question for clinical deployment is whether
task-specific fine-tuning suppresses L8H10's
annotation-reading behaviour. We tested two regimes.

\paragraph{Partial fine-tuning.} We updated only the last
two transformer layers and classification head for 5 epochs
(AdamW, lr=2e-5). The spurious rate changed from 61.5\% to
61.0\%, a reduction of one image in 200. Of 123 originally
spurious images, 122 (99.2\%) remained spurious.

\paragraph{Full fine-tuning.} We updated all 12 transformer
layers for 15 epochs using layerwise learning rate decay
(base lr=2e-5, decay factor 0.85 per layer toward input),
cosine annealing, and gradient clipping at 1.0. The spurious rate
was 62.5\%, marginally \emph{higher} than pretrained. Of 123
originally spurious images, 121 (98.4\%) remained spurious
after full fine-tuning. The mean corner/lung ratio declined
from 35.6$\times$ to 29.2$\times$ (Wilcoxon $p{=}0.009$),
but this shift is practically negligible: a ratio of
29$\times$ represents the same catastrophic
annotation-dominance as 36$\times$.

\begin{table}[t]
\centering
\caption{L8H10 spurious behaviour across fine-tuning regimes.}
\label{tab:finetuning}
\small
\begin{tabular}{@{}lccc@{}}
\toprule
\textbf{Condition} & \textbf{Spurious rate} &
\textbf{Mean ratio} & \textbf{Still spurious} \\
\midrule
Pretrained            & 61.5\% & 35.6$\times$ & ---    \\
Partial FT (2 layers) & 61.0\% & 34.8$\times$ & 99.2\% \\
Full FT (all layers)  & 62.5\% & 29.2$\times$ & 98.4\% \\
\midrule
\multicolumn{4}{l}{Full FT Wilcoxon $p{=}0.009$ vs pretrained} \\
\bottomrule
\end{tabular}
\end{table}

The fine-tuning objective does not provide a gradient signal that
contradicts corner-reading, because annotation text is correlated
with institutional disease prevalence in the training data. The
shortcut remains predictive throughout fine-tuning and is therefore
never penalised by the loss function. The slight increase in
spurious rate under full fine-tuning (61.5\%~$\to$~62.5\%) is
consistent: full fine-tuning strengthens task-relevant
representations, and because corner-reading is task-relevant in a
dataset with consistent annotation conventions, it is reinforced
rather than suppressed.
The practical implication is direct: \emph{the standard ViT
deployment pipeline, pretraining on ImageNet and fine-tuning on locally
labelled clinical data, does not remove spurious
annotation-reading and may entrench it.} A practitioner who
fine-tunes a pretrained ViT on chest X-rays from their institution
and then transfers the model to another site should not assume that
fine-tuning has cleaned up shortcut behaviour. Our results show it
has almost certainly preserved it, and may have made it worse.
Ablation-based mitigation strategies are structurally
insufficient: the spurious signal is distributed across redundant
components that collectively resist targeted removal. The only
known reliable mitigations are data-level, namely training on
annotation-stripped images so the shortcut is never learned, or
training-time regularisation explicitly targeting identified
spurious heads. Both require first knowing which heads to target.
This is exactly what mechanistic auditing provides.

%% ---------------------------------------------------------------
\section{Discussion}

\subsection{Structural Blindness in Output-Level Evaluation}
\label{sec:structural}

The null ablation result is a specific, falsifiable claim about
the structure of how neural networks encode information, with
direct implications for how AI safety evaluation is currently
conducted.

Elhage et al.~\cite{elhage2022toy} showed that Transformers
operating near capacity encode features in superposition: each
feature is distributed across many components. This creates a
fundamental asymmetry between what mechanistic analysis can detect
and what output-level evaluation can detect.

Output-level evaluation (accuracy, confidence, prediction
change rate) measures the \emph{aggregate} behaviour of the
network. It is blind to individual component behaviour unless
that component is both non-redundant and causally upstream of
the output. In a network with distributed representations,
neither condition holds for most components. A head can
implement a clearly spurious computation yet produce no output-level
signal when ablated, because the network reconstructs the same
output from the remaining redundant encoding.

The danger is not that L8H10 will cause the classifier to fail
on a single image. The danger is that the entire family of
spurious annotation-reading components constitutes a latent
vulnerability that activates silently under distributional shift.
When a model trained at one institution is deployed at another
with different annotation conventions, this distributed spurious
signal may suddenly dominate, and no output-level evaluation
performed at the training site will have detected it.

This is particularly consequential for ViT-based deployment
in low-resource clinical settings~\cite{vanitha2025vit},
where models are routinely transferred from high-resource
pretraining environments to new institutions without
re-auditing. The annotation conventions at the source
institution are encoded in the model's weights and will
transfer silently~--- invisible to any accuracy benchmark
evaluated on source-domain data.

\paragraph{What this means for evaluation practice.}
Current deployment evaluation for medical AI relies almost
entirely on output-level metrics: accuracy on held-out data,
performance under distribution shift, occasionally gradient
attribution maps. Our results demonstrate that this evaluation
regime is \emph{structurally blind} to a specific and dangerous
failure class: components that implement shortcuts at the
representational level but whose contribution is masked by
redundant encoding at the output level. This is not a
measurement problem that better ablation methods can solve;
it is a consequence of the architecture itself.
Standard evaluation answers: \emph{does the model perform
correctly on this data?} Mechanistic auditing answers:
\emph{which components are responsible, and are any doing so
for the wrong reasons?} These questions are not interchangeable,
and our results show a model can answer the first correctly while
failing the second catastrophically.
Mechanistic auditing of attention concentration, identification
of spurious-signal carriers, and causal verification of their
behaviour should be standard components of any pre-deployment
audit for Vision Transformers applied to medical imaging,
especially when models cross institutional boundaries.
\subsection{Mechanistic Interpretability Transfers to Vision Models}

Entropy analysis, attention distance metrics, and activation
patching developed for language models apply directly to ViT
attention heads with minimal modification. The abrupt
specialisation emergence in layers 7--9 mirrors functional circuit
emergence at intermediate depths in language models, suggesting the
phenomenon is architecture-general. Whether it extends to Swin,
BEiT, other medical modalities, or other pretraining regimes is an
open question.

\subsection{Limitations}

The corner region definition (top-left $3{\times}3$ patches)
and spurious threshold (corner/lung ratio ${>}2$) were chosen
to match the annotation convention of the datasets used here.
Datasets where annotations appear at other locations would require
adjusting the region definition accordingly. The underlying method
generalises directly: head identification depends only on max
single-patch attention concentration and mean attention distance,
neither of which assumes a specific annotation location.

Fine-tuning experiments used 200 images, which is small
relative to clinical training sets. The theoretical
argument for why suppression fails applies regardless of dataset
size when annotation conventions are consistent. The definitive
experiment would compare fine-tuning on annotation-consistent
versus annotation-stripped datasets of matched size; we
identify this as the most important direction for future work.

\subsection{Future Work}

\textit{Fine-tuning on annotation-stripped images} would directly
test whether suppressing the spurious signal during training
prevents it from being encoded. \textit{Automated spurious head
detection}, flagging heads with anomalously high max-attention
concentration as a pre-deployment screening step, would make
this analysis practical without per-model manual inspection.
\textit{Extension to other modalities} (retinal imaging,
dermoscopy, CT) would establish whether the phenomenon is
specific to chest X-rays or broader.

%% ---------------------------------------------------------------
\section{Conclusion}

We identified a single attention head, L8H10 in ViT-Base/16,
as a dedicated corner-contrast detector that attends to annotation
text in 62\% of clinical chest X-rays, with a corner-to-lung
attention ratio 74.8$\times$ higher in affected images than
unaffected ones ($p{<}0.0001$). Two causal interventions confirm
annotation content drives the behaviour. Replication on NIH
ChestX-ray14 (79\% rate, 37.9$\times$ separation) and across
DeiT-Base and ViT-Large/16 confirm generalisation across datasets
and architectures, with all three ViT-family models developing spurious
annotation-attending heads at majority rates.

Single-head ablation produces $p{=}0.340$. Multi-head ablation of
all three identified spurious heads produces $p{=}0.532$. The
spurious signal is distributed so broadly that even targeted
removal of all identifiable carriers leaves output-level evaluation
blind. A causally verified annotation-reading circuit is completely
invisible to standard evaluation. Mechanistic auditing finds it.
Behavioural testing does not. For Vision Transformers transferred
from natural image pretraining to clinical settings, including
the growing number deployed across resource-limited healthcare
systems, that gap warrants serious attention.

%% ---------------------------------------------------------------
\bibliographystyle{named}

%% ---------------------------------------------------------------
\appendix

\section*{Ethical Statement}

This study uses only publicly available, de-identified chest X-ray
datasets (COVID-19 Chest X-ray Dataset and NIH ChestX-ray14). No
new patient data were collected. All datasets were used in
accordance with their respective licences and intended research
purposes. The work identifies safety risks in deployed medical AI
systems and does not introduce any new harms.

\section*{Acknowledgments}

Anonymised for review.

\section{Experimental Setup}

All experiments on Google Colab (T4 GPU). Model:
\texttt{google/vit-base-patch16-224} via HuggingFace Transformers
v4.x. Primary dataset: \texttt{covid-chestxray-dataset}
(\url{github.com/ieee8023/covid-chestxray-dataset}). NIH
replication: \texttt{BahaaEldin0/NIH-Chest-Xray-14} via
HuggingFace Datasets. Classification model:
\texttt{codewithdark/vit-chest-xray} (CheXpert). Libraries:
PyTorch, NumPy, Matplotlib, SciPy, OpenCV. Code:
\url{https://github.com/joshmuhu/spurious-attention-vit}.

\section{Full Head Statistics}

Complete per-head entropy, distance, max attention, and focus
classification for all 144 heads in
\texttt{supplementary\_results.json}. Key heads: L7H2 (entropy
2.97, distance 2.00, max 0.094), L8H10 (entropy 2.88, distance
6.18, max 0.346), L9H3 (entropy 2.91, distance 2.28, max 0.090).
L8H10 is the only head combining low entropy with high spatial
distance and anomalously high single-patch concentration. The
next-highest max attention across all 144 heads is 0.124 (L6H12),
making L8H10 2.8$\times$ more concentrated than any other head in
the network.

\end{document}